\documentclass[conference]{IEEEtran}
\IEEEoverridecommandlockouts

\usepackage{cite}
\usepackage{amsmath,amssymb}
\usepackage{graphicx}
\usepackage{booktabs}
\usepackage{xcolor}
\usepackage{tikz}
\usetikzlibrary{arrows.meta,positioning,fit,backgrounds,calc}
\usepackage{algorithm}
\usepackage{algpseudocode}
\usepackage{listings}
\usepackage[hidelinks,breaklinks=true]{hyperref}
\usepackage{url}

\newcommand{\segdur}{T}
\newcommand{\code}[1]{\texttt{\small #1}}

\newcommand\blfootnote[1]{%
  \begingroup
  \renewcommand\thefootnote{}\footnote{#1}%
  \addtocounter{footnote}{-1}%
  \endgroup
}
\makeatletter
\g@addto@macro\UrlBreaks{\do\-\do\_\do\.}
\makeatother

\begin{document}

\title{Content-Based Video Narration of Gameplay\\
with Vision--Language Models}

\author{\IEEEauthorblockN{Mathew Varghese}
\IEEEauthorblockA{\textit{University of Washington} \\
Seattle, WA, USA \\
mathewvarghesemanu@gmail.com}}

\maketitle

\blfootnote{Work conducted independently; affiliation is listed for
identification only and does not imply institutional sponsorship or review.}
\blfootnote{Implementation available at
\url{https://github.com/mathewvarghesemanu/Content-based-video-narration-using-deep-learning}}

\begin{abstract}
Live game commentary is scarce: it exists for professional esports broadcasts
and almost nowhere else. We present a content-based video narration system that
produces spoken, esports-style commentary for arbitrary gameplay recordings
using a general-purpose vision--language model (VLM) and a text-to-speech back
end, with no game-specific instrumentation, no engine telemetry, and no
task-specific training. Three mechanisms carry the system. \emph{Temporal mosaic
packing} arranges nine uniformly sampled frames into a single $3\times3$ image,
letting an image-native VLM reason about motion while consuming one image
payload per segment instead of nine. \emph{Context-conditioned prompting}
replays the $K$ most recent narrations as assistant-role history, suppressing
the repetition that dominates per-segment captioning of static scenes.
\emph{Duration-conditioned generation and elastic alignment} constrain narration
length in the prompt, then time-scale or symmetrically pad the synthesized audio
so each utterance fills its segment slot exactly, giving frame-accurate muxing
without a forced aligner. The implementation supports either cloud TTS or a
6-bit quantized 4B-parameter on-device TTS model on Apple silicon, making the
speech stage fully local. We report a qualitative case study on real-time
strategy footage, a cost model showing the mosaic reduces per-minute image
payloads by $9\times$, and a candid account of observed failure modes:
hallucinated game state, resolution loss from mosaicking, and prosody artifacts
from time-scaling. We release the system as a reproducible baseline, with an
evaluation protocol for the quantitative study a full version will report.
\end{abstract}


\section{Introduction}

Commentary is part of how games are consumed. A professional esports broadcast
pairs the video feed with one or two human casters who name what is on screen,
attach stakes to it, and modulate their delivery to match the tension of the
moment. That layer is what turns a screen recording into something watchable by
someone who is not the player. It is also almost entirely absent outside the
professional tier: the overwhelming majority of gameplay video (amateur
uploads, training footage, replays, playtest captures, accessibility-oriented
recordings) ships silent or with only game audio.

The obvious reason is cost. Human commentary does not scale to the volume of
gameplay video produced, and the classical automation route does not help,
because it is not general. Prior systems that generate game commentary are
typically wired into a specific title: they read structured game state from an
engine, a replay file, or a telemetry API, and map that state to language with
templates or a trained
generator~\cite{ishigaki2021racing,harrison2017toward}. Such systems produce
accurate, well-grounded text, and they are useless for the next game, because
the state schema changes.

Modern vision--language models suggest a different trade. A general-purpose VLM
consumes pixels, which every game emits, and produces language, which is what
commentary is made of. Accuracy is traded away (the model does not know the
rules, cannot read a resource counter reliably, and will invent plausible
detail), but generality is bought: the same pipeline runs on a real-time
strategy replay, a first-person shooter clip, and a driving game without a line
of game-specific code. For the large space of gameplay video that has no
commentary at all, that trade is often the correct one.

This paper works through what it actually takes to make that trade concrete. The
naive construction (caption every frame, speak every caption) fails on
three separate axes, and each failure has a cheap fix that we adopt as a
contribution.

\begin{enumerate}
  \item \textbf{Cost and temporality.} Image-native VLMs charge per image and,
    when fed frames independently, see no motion. We pack a window of nine
    uniformly sampled frames into one $3\times3$ mosaic image
    (Section~\ref{sec:mosaic}), so a single request carries the temporal
    evidence of a whole segment at one image's cost. This follows the
    image-grid observation of Kim et al.~\cite{kim2024iggrid}, applied here
    under a hard latency and rate-limit budget.
  \item \textbf{Repetition.} Consecutive segments of gameplay look alike, and a
    stateless captioner responds with near-identical text, which is intolerable
    in speech where the listener cannot skim. We condition each request on the
    last $K$ generated narrations, injected as assistant-role turns
    (Section~\ref{sec:context}), and instruct the model not to repeat itself.
  \item \textbf{Synchronization.} Narration must fit the segment it describes,
    but generative text has unbounded length and TTS duration is not known
    until synthesis. We constrain length in the prompt with an explicit word and
    duration budget, then close the residual gap with elastic alignment:
    time-scaling long clips and symmetrically padding short ones
    (Section~\ref{sec:align}), yielding exact slot occupancy without
    forced alignment.
\end{enumerate}

We further show that the speech stage need not be a cloud dependency: a 6-bit
quantized 4B-parameter TTS model runs locally on consumer Apple silicon and
drops into the pipeline behind the same interface as the hosted TTS service,
which matters for cost, for offline use, and for footage that a user does not
want to upload.

Our contributions are: (i) a complete, reproducible, training-free pipeline for
spoken content-based gameplay narration, with cloud and on-device speech
back ends; (ii) the three mechanisms above, each isolated and described well
enough to reimplement; (iii) an analytical cost model for mosaic packing and a
qualitative case study on strategy-game footage; and (iv) an explicit account of
failure modes plus a pre-registered evaluation protocol for the quantitative
study we intend to report in the full-length version of this work. We position
this paper as a preprint baseline: the system and its mechanisms are complete
and reproducible, the human-subject evaluation is not yet run, and we state
which is which.

\section{Related Work}

\subsection{Video captioning and dense event description}
Describing video in natural language is a long-standing task, from
clip-level captioning on MSR-VTT~\cite{xu2016msrvtt} to dense event captioning,
which localizes and describes multiple overlapping events in untrimmed
video~\cite{krishna2017dense,zhou2018endtoend,yang2023vid2seq}. Gameplay
narration is dense captioning with two extra constraints that the benchmark task
does not impose: the output must be \emph{continuous} (every second of video
gets audio, not just interesting ones) and it must be \emph{speakable} within a
fixed time budget. Our elastic alignment stage exists precisely because the
second constraint has no analogue in text-output captioning.

\subsection{Sports and esports commentary generation}
Sports broadcasting is the closest applied neighbour. SoccerNet-Caption
introduced dense video captioning for soccer
broadcasts~\cite{mkhallati2023soccernet}, and GOAL added knowledge-grounded
commentary generation~\cite{qi2023goal}. On the games side, Ishigaki et
al.~generate racing-game commentary from vision, language, and structured
data~\cite{ishigaki2021racing}, and earlier work explores commentary and
explanation for game-playing agents~\cite{harrison2017toward}. The distinguishing
assumption in nearly all of this work is privileged access to structured state:
event logs, tracking data, or engine telemetry. We deliberately give that up.
Our system sees only what a viewer sees, which is what allows it to be pointed at
an arbitrary title.

\subsection{Vision--language models for video}
Instruction-tuned VLMs~\cite{liu2023llava,li2023blip2,alayrac2022flamingo,openai2023gpt4}
made general visual description practical without task-specific training, and a
line of work adapts them to video by sampling and projecting frame
features~\cite{maaz2024videochatgpt,zhang2023videollama,yang2023vid2seq}. Most
relevant to us, IG-VLM~\cite{kim2024iggrid} shows that simply arranging sampled
frames into an image grid lets an \emph{image} VLM perform competitively on
video question answering. We adopt that representation and report on its
practical costs in a production-shaped setting: the resolution loss it inflicts
on 1080p source frames is, in our experience, the dominant accuracy limitation of
the whole pipeline (Section~\ref{sec:limits}).

\subsection{Speech synthesis}
Neural TTS is mature enough to be treated as a component~\cite{shen2018tacotron2,kim2021vits},
and recent large TTS models produce expressive, controllable speech~\cite{wang2023valle}.
Our interest is narrower and systems-oriented: whether the speech stage can be
made local. We show it can, using a 6-bit quantized 4B model executed through an
Apple-silicon inference runtime, at quality we found acceptable for commentary
and with no per-request cost.

\subsection{Accessibility}
Automatic audio description for blind and low-vision audiences is an established
accessibility practice for film and television, and games remain poorly served by
it. Continuous narration of gameplay is structurally the same artifact as audio
description, produced under a different framing; we return to this in
Section~\ref{sec:ethics}.

\section{System}
\label{sec:system}

\subsection{Overview and notation}

The input is a video $V$ of duration $D$ seconds. We partition it into
$N = \lfloor D / \segdur \rfloor$ contiguous segments of $\segdur$ seconds. Each
segment $i$ receives exactly one narration utterance $u_i$, synthesized to an
audio clip $a_i$ whose duration is forced to a target $\tau$. The output is $V$
with its audio track replaced by $\bigoplus_{i} a_i$.

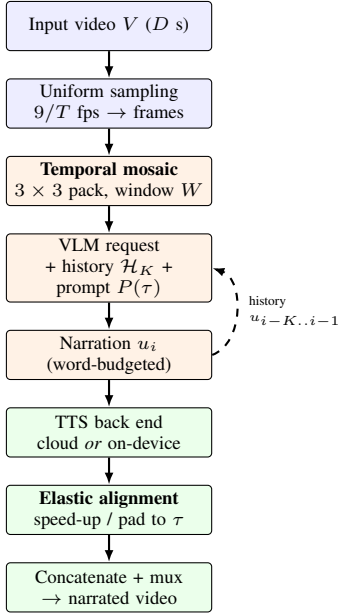
\begin{figure}[t]
\centering
\begin{tikzpicture}[
  node distance=3.6mm,
  every node/.style={font=\scriptsize},
  box/.style={draw, rounded corners=1.5pt, align=center, minimum height=6.5mm,
              text width=26mm, inner sep=1.6pt},
  vbox/.style={box, fill=blue!7},
  mbox/.style={box, fill=orange!10},
  abox/.style={box, fill=green!8},
  arr/.style={-{Latex[length=1.6mm]}, thick}
]
\node[vbox] (vid) {Input video $V$ ($D$ s)};
\node[vbox, below=of vid] (frames) {Uniform sampling\\$9/\segdur$ fps $\rightarrow$ frames};
\node[mbox, below=of frames] (mosaic) {\textbf{Temporal mosaic}\\$3\times3$ pack, window $W$};
\node[mbox, below=of mosaic] (vlm) {VLM request\\+ history $\mathcal{H}_{K}$ + prompt $P(\tau)$};
\node[mbox, below=of vlm] (text) {Narration $u_i$\\(word-budgeted)};
\node[abox, below=of text] (tts) {TTS back end\\cloud \emph{or} on-device};
\node[abox, below=of tts] (scale) {\textbf{Elastic alignment}\\speed-up / pad to $\tau$};
\node[abox, below=of scale] (mux) {Concatenate + mux\\$\rightarrow$ narrated video};

\foreach \a/\b in {vid/frames, frames/mosaic, mosaic/vlm, vlm/text, text/tts,
                   tts/scale, scale/mux}
  \draw[arr] (\a) -- (\b);

\draw[arr, dashed] (text.east) to[out=20,in=-20]
  node[midway, right=0.4mm, font=\tiny, align=left] {history\\$u_{i-K..i-1}$} (vlm.east);
\end{tikzpicture}
\caption{Pipeline. Blue: video decoding. Orange: language generation. Green:
speech and assembly. The dashed edge is the context-conditioning loop that
suppresses cross-segment repetition.}
\label{fig:pipeline}
\end{figure}

Frames are sampled uniformly at $9/\segdur$ frames per second, so that each
segment contributes nine frames, exactly the capacity of a $3\times3$ mosaic.
This coupling of the sampling rate to the mosaic geometry is deliberate: it makes
$\segdur$ the single knob that trades temporal resolution against request count.
Fig.~\ref{fig:pipeline} shows the full pipeline.

\subsection{Temporal mosaic packing}
\label{sec:mosaic}

For segment $i$ we take a window $W$ of consecutive sampled frames centred on
the segment (with clamping at the sequence boundaries), and composite them into a
single image on a $3\times3$ lattice in raster order, so that reading order
matches time order:
\[
M_i[x + w\!\cdot\!c,\; y + h\!\cdot\!r] = f_{9i + 3r + c}[x, y],
\quad r,c \in \{0,1,2\},
\]
where $w \times h$ is the source frame size. The composite is JPEG-encoded,
base64-embedded, and sent as one image in the request.

The motivation is economic as much as representational. A hosted VLM prices and
rate-limits per image; the deployment we developed against imposed a hard daily
request cap, which for a naive per-frame pipeline is exhausted by a few minutes
of footage. Let $R$ be the images consumed per minute of video. Per-frame
captioning at the same temporal resolution gives $R_{\text{frame}} = 540/\segdur$
images per minute, whereas mosaic packing gives $R_{\text{mosaic}} = 60/\segdur$,
a $9\times$ reduction, independent of $\segdur$. At our default
$\segdur = 9$\,s this is $6.7$ images per minute rather than $60$, which is the
difference between narrating an hour of footage inside a daily cap and not.

The representational argument is that the mosaic exposes change. A single frame
cannot show that a unit moved; nine tiles in reading order can, and the model
does describe motion and escalation given them (Section~\ref{sec:case}). The
cost, which we do not minimize, is spatial: nine 1080p tiles form a
$5760\times3240$ composite that the API downsamples aggressively, so fine detail
(resource counters, health bars, minimap icons, small text) is effectively
destroyed, while large overlay text (alerts, banners) generally survives. This
asymmetry is the root cause of most content errors we observed
(Section~\ref{sec:case}).

\subsection{Context-conditioned prompting}
\label{sec:context}

Gameplay is visually autocorrelated: adjacent segments of a base-building phase
or a positional standoff look nearly identical. A stateless captioner therefore
returns nearly identical text, and repetition that is merely dull in subtitles
is unbearable in speech.

We condition generation on recent output. Let $\mathcal{H}_K =
\langle u_{i-K}, \dots, u_{i-1}\rangle$ be the last $K$ narrations. These are
inserted into the request as \emph{assistant}-role turns preceding the current
user turn, so the model perceives them as its own prior utterances rather than as
instructions, and the task prompt closes with an explicit directive to maintain
context and avoid repeating previous messages. We use $K=5$ by default; larger
$K$ increases prompt tokens linearly and, in our observation, yields diminishing
novelty benefit once the window covers the visible scene phase.

This mechanism does double duty. Besides suppressing repetition, it produces
narrative continuity: because the model can see that it already said an attack
was building, it tends to describe the next segment as escalation or resolution
rather than as a fresh scene. That is a large part of what makes the output sound
like commentary rather than like a sequence of captions.

\subsection{Duration-conditioned generation}
\label{sec:duration}

Each utterance must be speakable in $\tau$ seconds. We set $\tau = 2\segdur$,
i.e.\ narration audio runs at twice the wall-clock rate of segment progression,
which reflects the design choice that commentary is continuous over the video
rather than sparse; $\tau$ is a free parameter and setting $\tau = \segdur$
yields exactly real-time narration.

The target is injected into the prompt by template substitution, together with a
hard word budget and a style directive:

\begin{lstlisting}
Quickly analyze the attached game screenshot.
Identify the most prominent features: key units,
any ongoing combat, notable buildings, and current
resources without going into math. Provide a brief
description focusing only on these critical aspects,
suitable for a quick {audio_duration}-second overview
less than 40 words like realtime e-sports commentary
with emotional prompts. Don't mention the name of the
game or the info that this is a screenshot. Keep in
mind the context and try not to repeat the previous
messages. Target audience is esports fans.
\end{lstlisting}

Three parts of this prompt are load-bearing and were each added in response to an
observed failure. The word budget ($<40$ words) bounds synthesis duration
enough that the alignment stage rarely has to apply aggressive time-scaling.
Suppressing the meta-frame (``don't mention that this is a screenshot'') removes
a persistent artifact in which the model narrated the medium rather than the
match, the single most immersion-breaking failure we encountered. Naming the
audience and register (``esports fans'', ``emotional prompts'') is what moves the
output from neutral description into commentary voice, and it is a substantial
part of the perceived quality difference.

We also cap generation with a token limit as a second line of defence, and we
detect refusal-shaped responses (a leading apology) and retry, up to three
attempts, since a spoken refusal in the middle of a narration track is a total
failure of the artifact.

\subsection{Speech synthesis with interchangeable back ends}
\label{sec:tts}

The TTS stage is behind a two-function interface
(\code{generate\_tts(text, path)}), with two implementations selected by
configuration:

\begin{itemize}
  \item \textbf{Cloud.} A hosted high-definition TTS model (\code{tts-1-hd})
    with a selectable voice. Highest quality and lowest engineering cost; incurs
    per-character charges and requires uploading the narration text.
  \item \textbf{On-device.} A 4B-parameter Voxtral TTS model quantized to 6 bits,
    executed on Apple silicon through the MLX array framework, exposing 19
    voices across several languages. Audio is generated as 24\,kHz float
    samples, concatenated across streaming chunks, and encoded to MP3. No API
    key, no network, no per-request cost.
\end{itemize}

That the on-device path is viable changes the deployment envelope. The VLM stage
still requires a hosted model, but the stage that scales with narration
\emph{volume}, and that would dominate cost on a long video, runs locally.
It also removes a privacy objection for unreleased or personal footage, at least
for the text-to-audio half of the pipeline.

\subsection{Elastic temporal alignment}
\label{sec:align}

Synthesized clip durations are not controllable a priori: the same 35-word
narration may render to 8 or 13 seconds depending on voice and phonetic content.
We therefore force each clip to exactly $\tau$ after the fact. Given clip $a_i$
of duration $d_i$:
\[
a_i' =
\begin{cases}
  \mathrm{speedup}\!\left(a_i,\ d_i/\tau\right), & d_i > \tau, \\[2pt]
  \mathrm{sil}\!\left(\tfrac{\tau - d_i}{2}\right) \oplus a_i \oplus
    \mathrm{sil}\!\left(\tfrac{\tau - d_i}{2}\right), & d_i \le \tau,
\end{cases}
\]
where $\mathrm{speedup}$ is phase-vocoder time-scaling that preserves pitch and
$\mathrm{sil}(t)$ is $t$ seconds of silence. Because every $a_i'$ has duration
exactly $\tau$, concatenation places utterance $i$ at a known offset $i\tau$ and
the final mux is a stream copy of the video with the new audio track: no
forced aligner, no drift accumulation, no re-encode of the video.

The symmetric padding is a small but consequential detail: centring the utterance
in its slot means the speech onset lags the segment boundary rather than
colliding with it, which reads as a caster reacting to what just happened
instead of pre-empting it. The failure mode of this stage is compression on the
other side: when the model overshoots its word budget, ratios above roughly
$1.3$ produce audibly rushed, chipmunked delivery. Tightening the word budget is
a better fix than tolerating the artifact.

\subsection{Algorithm}

\begin{algorithm}[t]
\caption{Content-based gameplay narration}
\label{alg:main}
\begin{algorithmic}[1]
\Require video $V$, segment length $\segdur$, history $K$, target $\tau=2\segdur$
\State $F \gets \textproc{SampleFrames}(V,\ 9/\segdur)$
\State $\mathcal{H} \gets \langle\,\rangle$;\quad $U \gets \langle\,\rangle$
\For{$i = 0$ \textbf{to} $\lfloor |F|/9 \rfloor - 1$}
  \State $M_i \gets \textproc{Mosaic3x3}(F,\ \text{window centred at } i)$
  \For{$\text{attempt} = 1$ \textbf{to} $3$}
    \State $u \gets \textproc{VLM}(P(\tau),\ M_i,\ \text{last } K \text{ of } \mathcal{H})$
    \If{$u$ valid \textbf{and not} refusal-shaped} \textbf{break} \EndIf
  \EndFor
  \State append $u$ to $\mathcal{H}$ and to $U$
\EndFor
\For{each $u_i \in U$}
  \State $a_i \gets \textproc{TTS}(u_i)$ \Comment{cloud or on-device}
  \State $a_i' \gets \textproc{FitToDuration}(a_i,\ \tau)$
\EndFor
\State \Return $\textproc{Mux}(V,\ \bigoplus_i a_i')$
\end{algorithmic}
\end{algorithm}

\section{Implementation}

The system is implemented in Python. Frame extraction and final muxing use
FFmpeg; mosaic composition uses Pillow; audio time-scaling, padding, and
concatenation use pydub. Every parameter in Table~\ref{tab:config} is exposed in
a single YAML file, so a run is fully described by that file plus the input
video.

\textbf{Models.} The vision stage uses OpenAI \emph{GPT-4o}, called through the
chat completions HTTP endpoint with the mosaic base64-inlined as a single image
and a 300-token completion cap. An earlier iteration of the system used
\code{gpt-4-vision-preview}; the pipeline is otherwise unchanged, which is weak
evidence that it is not tied to a particular model generation. The cloud speech
stage uses OpenAI \code{tts-1-hd} with the \emph{nova} voice. The on-device
speech stage uses a 6-bit quantized 4B-parameter Voxtral TTS checkpoint executed
through MLX with \code{soundfile} for audio I/O.

None of the three mechanisms in Section~\ref{sec:system} depends on this choice.
Mosaic packing requires only that the model accept an image; context conditioning
requires only a chat-style role-tagged history; duration conditioning is prompt
text. Any instruction-following VLM with an image input and any TTS with a file
output can be substituted, and we name the specific models here because the
qualitative results in Section~\ref{sec:case} are not reproducible without
them, not because the design assumes them.

\begin{table}[t]
\caption{Configuration surface and defaults}
\label{tab:config}
\centering
\footnotesize
\begin{tabular}{@{}llp{0.31\columnwidth}@{}}
\toprule
\textbf{Parameter} & \textbf{Default} & \textbf{Effect} \\
\midrule
\code{segment\_duration} $\segdur$ & 9\,s & sampling rate, request count \\
\code{vision.window\_size} & 9 & frames per mosaic \\
\code{vision.history\_size} $K$ & 5 & repetition suppression \\
\code{vision.max\_tokens} & 300 & bound on utterance length \\
\code{vision.prompt} & \S\ref{sec:duration} & register, budget, grounding \\
\code{tts.method} & on-device & cloud vs.\ local speech \\
target duration $\tau$ & $2\segdur$ & narration density \\
\bottomrule
\end{tabular}
\end{table}

Two engineering details matter for reproducibility. Frame extraction is
idempotent (it is skipped when frames already exist) because decoding
dominates wall-clock time on long inputs and iterating on prompts should not pay
for it repeatedly. Previous narration transcripts are archived with a timestamp
rather than overwritten, since prompt iteration is the main development loop and
comparing successive transcripts is how one evaluates a prompt change.

\section{Qualitative Case Study}
\label{sec:case}

We ran the pipeline on a 60-second, $1920\times1080$, 60\,fps capture of an
\emph{Age of Empires II: Definitive Edition} match (Castle Age, eight players,
an active raid in progress), with $\segdur = 9$\,s, $K = 5$, $\tau = 18$\,s,
and mosaic window $W = 9$, using GPT-4o for narration. Table~\ref{tab:output}
gives the verbatim output of the first three segments, unedited.

\begin{table}[t]
\caption{Verbatim narration, RTS footage, $\segdur=9$\,s}
\label{tab:output}
\centering
\footnotesize
\begin{tabular}{@{}p{0.06\columnwidth}p{0.86\columnwidth}@{}}
\toprule
\textbf{Seg} & \textbf{Generated narration} \\
\midrule
0 & ``Red alert! Villagers under heavy attack by Lorraine's forces! Castle
stands strong at the heart, surrounded by buzzing farms. Resources are holding,
but pressure is mounting. Will they withstand this relentless assault?'' \\
\addlinespace[2pt]
1 & ``Amidst the chaos, Lorraine and Burgundy strike! Cavalry charges fiercely!
Villagers scurry, farms buzz with productivity. The castle stands resolute,
resources steady. Can this defense hold strong? The tension's palpable!'' \\
\addlinespace[2pt]
2 & ``Tension peaks! Cavalry units clash fiercely near the crucial farms!
Villagers hustle, castles and buildings stand firm. Resources hold, but Burgundy
and Lorraine's aggressive push intensifies! Defense must rally, or the momentum
could shift dramatically!'' \\
\bottomrule
\end{tabular}
\end{table}

Several properties are visible in this sample and are, in our experience,
representative.

\textbf{Entity grounding works at the coarse level.} Villagers, cavalry, farms,
and the castle are correctly identified from appearance alone. The opponent names
(Lorraine, Burgundy) are also correct, but their provenance is worth separating:
the footage carries an in-game alert overlay reading ``You are being attacked by
5 Lorraine'' and ``7 Burgundy'', so the model is reading large overlay text
rather than recognizing civilizations visually. Both capabilities are useful and
neither required game-specific configuration, which is the premise of the
approach, but only the first is visual recognition, and a paper that conflated
them would overstate what the VLM is doing.

\textbf{Register transfer works.} The output is unmistakably in commentary voice:
second-person stakes, rhetorical questions, escalation vocabulary. This comes
entirely from the audience-and-register clause of the prompt.

\textbf{Continuity emerges from the history mechanism.} The three segments form
an arc (assault begins, defenders hold, tension peaks) rather than three
independent descriptions of a similar scene. Nothing in the system models the
match state; the arc is a side effect of the model seeing its own prior
utterances. Lexical repetition is nonetheless still visible (``farms'',
``resources hold'', ``stands strong'' recur), which tells us $K=5$ history
attenuates but does not eliminate the problem.

\textbf{Confidence outruns evidence.} ``Resources are holding'' and ``resources
steady'' appear in all three segments. The source frames in fact carry an exact,
legible resource bar (1125 wood, 657 food, 400 gold, 894 stone, 35/70
population); after $3\times$ spatial reduction in the mosaic and further
downsampling by the API, those digits are a few pixels tall and unreadable. The
model is therefore emitting a plausible commentary phrase, not a reading of game
state, and the information it is bluffing about was present in the input and
discarded by our own representation. This is the cleanest available illustration
of the cost side of mosaic packing, and it is not hedged: a listener could not
identify it as invention. We treat it as the central open problem of the approach
and discuss it below.

\textbf{Prompt--content mismatch is tolerated.} The prompt used for this run was
left over from an earlier experiment and explicitly named a popular first-person
shooter title, while the footage was real-time strategy. The model ignored the
stated title and described what it actually saw,
in genre-appropriate vocabulary. This is accidental but useful evidence for
generality (pixels dominate the prompt's genre hint), and it simultaneously
shows that the prompt cannot be relied upon to steer the model into a specific
title's domain vocabulary when that is what one wants.

\subsection{Cost}
At $\segdur = 9$\,s the pipeline issues $6.7$ VLM requests and $6.7$ TTS
utterances per minute of video, against $60$ VLM images per minute for
per-frame captioning at identical temporal resolution. With the on-device speech
back end the TTS component of marginal cost is zero and the pipeline's monetary
cost is exactly the VLM requests.

\section{Proposed Evaluation Protocol}
\label{sec:eval}

This preprint reports the system and a qualitative case study; it does not report
a human-subject evaluation. We state the protocol we intend to run so that the
claims of the full version are pre-registered rather than retrofitted.

\textbf{Corpus.} 30 clips of 60--120\,s spanning three genres (RTS, FPS, racing),
10 per genre, drawn from publicly available gameplay recordings.

\textbf{Conditions.} (C1) full system; (C2) no history ($K = 0$), isolating
context conditioning; (C3) per-frame captioning instead of mosaic, at matched
temporal resolution and $9\times$ the request cost, isolating mosaic packing;
(C4) human-written commentary on a subset, as a ceiling; (C5) original game audio
only, as a floor.

\textbf{Subjective measures.} Mean opinion score on a 5-point scale for
naturalness, informativeness, excitement/appropriateness of register, and
synchronization, collected from participants who play the relevant genre, with
each clip rated by at least five raters and inter-rater agreement reported.

\textbf{Objective measures.} (i) \emph{Factual grounding rate}: fraction of
verifiable claims per utterance that are correct, annotated against the video by
two independent annotators. This is the metric that directly targets the
hallucination failure above. (ii) \emph{Repetition}: mean pairwise ROUGE-L and
embedding cosine similarity between consecutive utterances, which should separate
C1 from C2 if context conditioning does what we claim. (iii) \emph{Alignment
error}: distribution of $|d_i - \tau|$ before correction, and the fraction of
clips requiring a time-scale ratio above 1.3. (iv) \emph{Cost and latency}:
requests, tokens, and wall-clock seconds per minute of video, cloud versus
on-device speech.

\textbf{Accessibility study.} A separate small-$n$ study with blind and
low-vision participants, evaluating the output as audio description rather than
as entertainment, since the two use cases weight informativeness and excitement
very differently and we expect the current prompt to be tuned wrongly for the
former.

\section{Limitations and Failure Modes}
\label{sec:limits}

\textbf{Hallucinated game state.} The model asserts quantities it cannot see.
Because the register is confident by construction, these assertions are
indistinguishable from grounded ones to a listener. Any use of this system where
a viewer might act on the commentary as information (coaching, analysis,
accessibility) requires this to be fixed, not merely noted. Two directions
apply: prompt-level hedging that permits the model to omit rather than invent,
and hybrid grounding that supplies verified values for it to use, which we
develop under the next item since the two problems share a cause.

\textbf{Resolution loss from mosaicking.} Nine 1080p tiles are downsampled to a
degree that destroys HUD text and small sprites. This is a direct trade against
the $9\times$ cost saving, and it is the mechanism behind the previous failure.

We believe the strongest available fix is \emph{region-of-interest
decomposition}, and we state it here as the first item of future work because it
attacks the hallucination and the resolution loss with one mechanism. Game
interfaces are not visually uniform: they are spatially structured, and the
structure is stable across an entire session. In the strategy footage of
Section~\ref{sec:case} the screen decomposes into a resource and population bar
(top-left), an objectives panel (top-right), an alert and chat overlay
(upper-left), a scoreboard (lower-right), a selected-unit and production panel
(bottom-centre), and a minimap (bottom-right), leaving the playfield as the
only region where the $3\times3$ mosaic's motion evidence actually matters. Each
non-playfield region is small, fixed in position, and information-dense in
exactly the way that survives poorly under downsampling.

This suggests a two-stream request rather than a single flattened image. The
playfield is mosaicked as now, at low resolution, because what is being asked of
it is coarse and temporal: what moved, what is fighting, where the pressure
is. The HUD regions are cropped at native resolution and handled separately and
cheaply: numeric fields (resources, population, score, timers) by OCR, the
minimap by colour-blob analysis giving contact locations and rough force
distribution, and alert and chat overlays by direct text extraction, which the
case study shows already works. The extracted values are then injected into the
prompt as a structured state block, and the prompt is constrained to use those
numbers and to say nothing quantitative that does not appear in them. This
converts the model's job from guessing at state to narrating supplied state,
which is where LLMs are reliable.

Two properties make this attractive beyond accuracy. The HUD readers are cheap
(OCR on a handful of small crops costs no VLM tokens), so the approach does
not give back the $9\times$ saving that motivated mosaicking. And the cost of
generality is bounded and honest: region coordinates are per-title configuration,
roughly a dozen rectangles in a YAML file, produced once per game and, for
titles with stable interfaces, reusable across patches. That is a far weaker
game-specific dependency than the engine or telemetry access that prior
commentary systems assume (Section~II), and it degrades gracefully: with no
region file the system behaves exactly as it does today. Auto-detecting HUD
regions by observing which screen areas remain static across many frames while
the playfield changes is a plausible way to remove even that step, and we have
not attempted it.

\textbf{No audio input.} The system never listens to the game. Gunfire, ability
sounds, and voice lines carry event information that the visual channel often
lacks, and ignoring them is a significant missed signal.

\textbf{Fixed segmentation.} Segments are uniform, so a decisive teamfight and an
idle economy phase receive equal narration budget. Event-driven segmentation
(detect salience, then allocate narration) is the obvious improvement and would
change the architecture more than any other item on this list.

\textbf{Batch, not live.} The pipeline is offline. Live use requires streaming
frame capture, latency-bounded generation, and incremental muxing; the per-request
round trip is the binding constraint.

\textbf{Single voice, no interplay.} Real broadcasts use play-by-play and colour
commentary in dialogue. We emit one voice; the multi-voice extension is
straightforward with the existing back ends and untried.

\textbf{Prosody artifacts.} Time-scaling beyond roughly $1.3\times$ is audible.
Tighter word budgets, or duration-aware regeneration, are preferable to
post-hoc compression.

\textbf{No quantitative evaluation yet.} As stated in Section~\ref{sec:eval}. The
claims in this paper about repetition suppression and register transfer are
qualitative observations from development, not measured effects.

\section{Ethical Considerations}
\label{sec:ethics}

\textbf{Accessibility.} The most defensible use of this system is audio
description of games for blind and low-vision players and viewers, a population
that commercial games serve poorly. That use case demands accuracy over
excitement, which inverts the current prompt's priorities, and it makes the
hallucination problem a safety-relevant defect rather than a quality one.

\textbf{Labour.} Automatic commentary touches the work of human casters. We note
that the realistic near-term application is footage that would otherwise have no
commentary at all (amateur uploads, replays, archives) rather than
professional broadcasts, where the quality gap is wide and the social function of
a known caster is not substitutable. We do not think this observation dissolves
the concern; we think it bounds it.

\textbf{Provenance.} Generated commentary should be labelled as such. Confident
synthetic narration over real footage is a plausible vector for misrepresenting
what happened in a match, and the system as built provides no watermark or
disclosure mechanism.

\textbf{Content and rights.} Gameplay footage is subject to the rights of both
the capturing player and the game publisher, and the cloud paths transmit
frames and text to third-party services. The on-device speech back end reduces
but does not remove that exposure.

\section{Conclusion}

We presented a training-free system that narrates arbitrary gameplay video with
spoken, esports-style commentary using a general vision--language model and an
interchangeable speech back end, one of which runs entirely on-device. Three
mechanisms carry the system: temporal mosaic packing, which buys motion evidence
at a ninth of the image cost; context-conditioned prompting, which converts
independent captions into a continuous narrative; and duration-conditioned
generation with elastic alignment, which makes speech fit video exactly without a
forced aligner. A qualitative case study on strategy-game footage shows correct
coarse entity grounding, successful register transfer, and emergent continuity,
alongside confident hallucination of game state that our own mosaic had
discarded. We identify that as the central unsolved problem, and region-of-interest
decomposition (reading HUD panels, minimap, and alert overlays at native
resolution while the playfield stays mosaicked) as the first target of future
work. We have
specified the evaluation protocol for the quantitative study rather than
reporting partial numbers, and we release the implementation as a reproducible
baseline for a task that, despite abundant gameplay video, has almost no
open tooling.

\section*{Acknowledgments}
The author thanks the open-source maintainers of FFmpeg, Pillow, pydub, and MLX,
without which this system would have been substantially more work than it was.

\bibliographystyle{IEEEtran}
\bibliography{refs}

\end{document}